\documentclass[final]{cvpr}
\usepackage{times}
\usepackage{epsfig}
\usepackage{graphicx}
\usepackage{amsmath}
\usepackage{amssymb}
\usepackage{caption}
\usepackage{amsmath} 
\usepackage{color}
\usepackage{booktabs}
\usepackage{amssymb}
\usepackage{pifont}
\newcommand{\cmark}{\ding{51}}%
\newcommand{\xmark}{\ding{55}}%
\newcommand{\bc}[1]{\textcolor{blue}{#1}}
\newcommand{\rc}[1]{\textcolor{red}{#1}}

\newcommand{\tabincell}[2]{\begin{tabular}{@{}#1@{}}#2\end{tabular}}
\usepackage{multirow}
\newcommand{\heading}[1]{\noindent\textbf{#1}}

\usepackage[pagebackref=true,breaklinks=true,colorlinks,bookmarks=false]{hyperref}

\def\cvprPaperID{4279} 
\def\confYear{CVPR 2021}

\begin{document}

\title{Image Inpainting Guided by Coherence Priors of Semantics and Textures}

\author{Liang Liao$^{1}$ \and Jing Xiao$^{2}$ \and
Zheng Wang$^{1}$ \and Chia-Wen Lin$^{3}$ \and Shin'ichi Satoh$^{1}$\\
\and
$^{1}$National Institute of Informatics \and $^{2}$Wuhan University \and  $^{3}$National Tsing Hua University\\
}



\maketitle

\begin{abstract}
Existing inpainting methods have achieved promising performance in recovering defected images of specific scenes. However, filling holes involving multiple semantic categories remains challenging due to the obscure semantic boundaries and the mixture of different semantic textures. In this paper, we introduce coherence priors between the semantics and textures which make it possible to concentrate on completing separate textures in a semantic-wise manner. Specifically, we adopt a multi-scale joint optimization framework to first model the coherence priors and then accordingly interleavingly optimize image inpainting and  semantic segmentation in a coarse-to-fine manner. A Semantic-Wise Attention Propagation (SWAP) module is 
devised to refine completed image textures across scales by exploring non-local semantic coherence, which effectively mitigates mix-up of textures. We also propose two coherence losses to constrain the consistency between the semantics and the inpainted image in terms of the overall structure and detailed textures. Experimental results demonstrate the superiority of our proposed method for challenging cases with complex holes. 
\end{abstract}

\section{Introduction}

High-quality image inpainting aims to fill in missing regions with synthetic content \cite{barnes2009patchmatch,bertalmio2000image,criminisi2004region}. It requires both semantically meaningful structures and visually pleasing textures. To this end, deep learning-based methods ~\cite{pathak2016context,yan2018shift,yu2018generative,AAAI-normlization,Zeng-highreso,8067496} resort to encoder-decoder based networks to infer the context of a corrupted image and then refine the texture details in the initial inference of a missing region by some tools, such as non-local algorithms. Although current image inpainting methods have made significant progress, it still poses technical challenges in completing complex holes, particularly when a missing region involves multiple sub-regions with different semantic classes. The main reason falls in the failure of modeling the prior distributions of a mixture of different semantic regions, which usually result in blurry boundaries and unrealistic textures \cite{Liao2020guidance}.

\begin{figure}[tb]
\centering
\includegraphics[width=\linewidth]{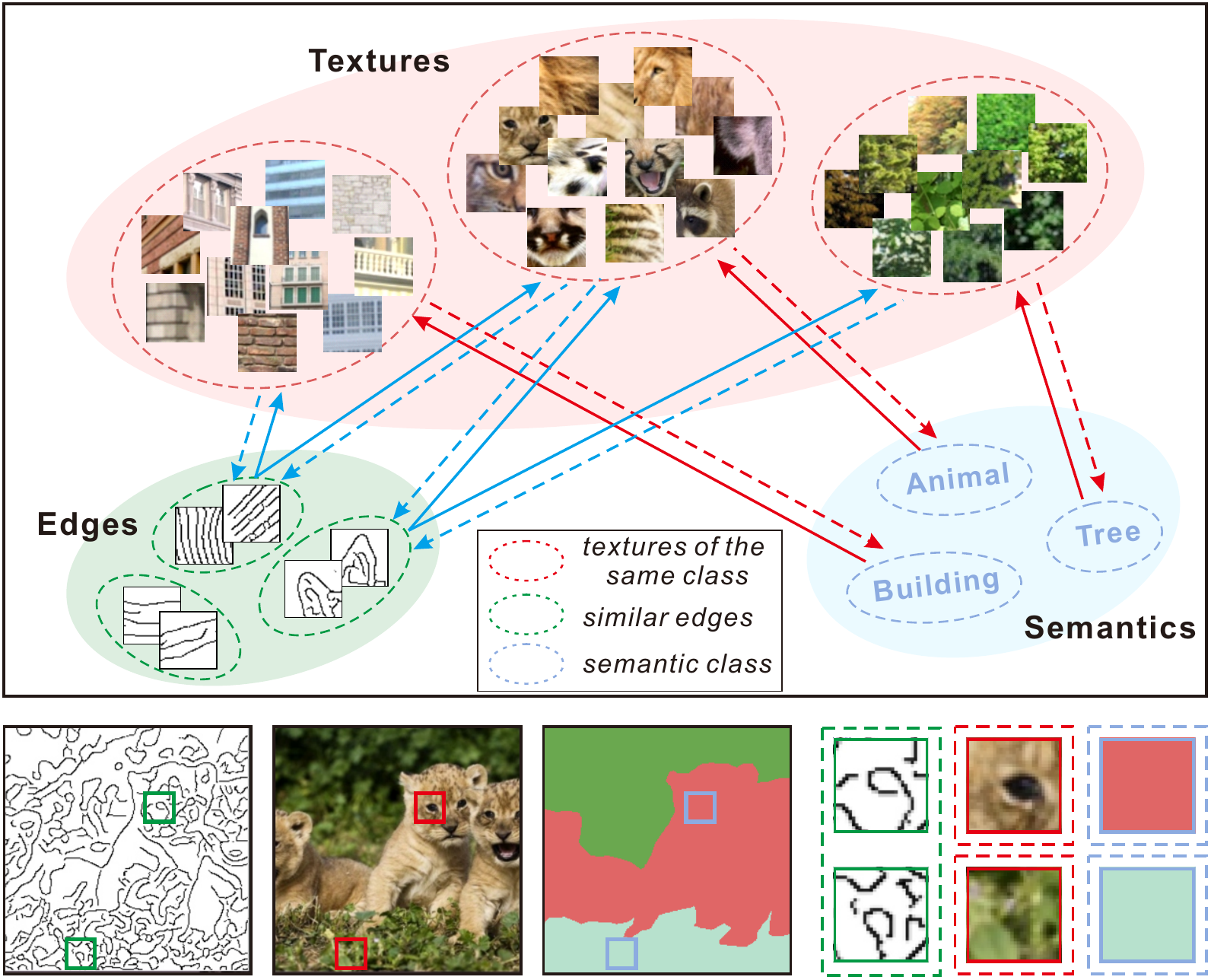}
\caption{\small Upper part: Mapping between image textures and edges/semantics (Dot arrow - extraction of edges/semantics; solid arrow - texture generation). Notice that two similar edge patches in a green circle could be mapped to completely different semantic textures, but one semantic will be clearly mapped to a certain texture category. Lower part: an example showing two similar edge patches map to two different semantic textures.}
\vspace{-2mm}
\label{fig:fig1}
\end{figure}

A feasible approach is to adopt structural information, such as edges \cite{li2019structure,nazeri2019edgeconnect}, contours \cite{xiong2019foreground}, and smooth images \cite{ren2019structureflow}, as guidance to complete missing structures and textures in two steps. The assumption is that structures offer semantic clues for inferring an unknown scene, making them suitable for guiding the filling of textures. However, we notice that the correspondence between structural information and textures is not apparent, making the filled textures still highly rely on the local correlation around the missing region. Figure~\ref{fig:fig1} demonstrates the ambiguity of the mapping from mid-level structures (e.g., the edges) to the textures, which can significantly degrade the visual authenticity of the generated textures.

Compared with mid-level structures, high-level semantic information offers more vital semantic clues to the object textures. For example, in Figure~\ref{fig:fig1}, the semantic \textit{animal} leads to fluffy fur while the semantic \textit{tree} leads to green leaf in the image, which cannot be usually distinguished solely from their mid-level structures. In contrast, object textures have been shown to provide sufficient information about the semantic classes \cite{Geirhos2019texture,KamannR20}. Thus we characterize the relationships between the semantics and textures of objects as \textbf{coherence priors} and build our inpainting method on the coherence priors to complete complex holes while ensuring the mutual consistency between the predicted semantics and textures.


Based on the above motivation, we propose to utilize coherence priors between semantics and textures to facilitate joint optimization of semantic segmentation and image inpainting. To this end, our framework extracts a shared feature to represent the common information of the two tasks, and characterize the interaction between scales to enable the utilization of coherence priors to optimize the two tasks jointly. Specifically, two novel designs are proposed:
1) A Semantic-Wise Attention Propagation (SWAP) module is used to explicitly capture the semantic relevance between an unknown (missing) area and the known regions. As a result, when mapping semantics to image textures, filling in an unknown patch only refers to those known patches with the same semantic, rather than to the entire image, to avoid irrelevant texture filling.
2) We devise two loss terms to learn the global and local coherence relationships, respectively. The image-level structure coherence loss is used to supervise the structural matching between the inpainted image and the corresponding segmentation map to generate clear boundaries in the inpainted image. Besides, the non-local patch-level coherence loss aims to assess the distribution of patch textures in the semantic domain to encourage the generated textures to be as similar as the matched known patch of the same semantics.

Unlike the existing semantics-guided inpainting methods, such as SPG-Net \cite{song2018spg} and SGE-Net \cite{Liao2020guidance}, which synthesize textures by convolution to involve the local semantic information, our proposed method predicts the textures and the semantics simultaneously, and borrows the known texture feature of the same semantic to fill in a missing region by a semantics-guided non-local means, which not only ensures realistic textures but also is valuable for semantic recognition. For a better understanding of our method, we summarized the comparison to other methods in Table~\ref{tab:relatedwork}. The main contributions of our paper are three-fold:

\begin{itemize}
    \item We introduce coherence priors that highlight the mutual consistency between the semantics and textures in image inpainting and devise two coherence losses to boost the consistency between semantic information and inpainted image in the global structure level and local texture level. 
    \item We propose a semantic-wise attention propagation module, which generates semantically realistic textures by capturing distant relationship and  referring to the texture feature of the same semantic in the feature maps.
    \item Our approach outperforms existing state-of-the-art image inpainting methods, including semantics-based approaches \cite{Liao2020guidance,nazeri2019edgeconnect,song2018spg,yu2019free,zeng2019learning} on completing complex hole with multiple semantic regions in terms of the sharpness of boundaries and the coherence and visual plausibility of textures. 
\end{itemize}
 
\section{Related Work}

\subsection{Image inpainting}

Deep learning-based inpainting approaches have recently been proposed by understanding the images, which can generate meaningful content for filling in the missing region. Context Encoder \cite{pathak2016context} was first proposed to employ a generative adversarial network and demonstrated its potential for inpainting tasks. Based on it, efforts have been made to enhance the inpainting performance, including introducing specific losses \cite{dosovitskiy2016generating,Wang-column,8067496}, building recursive architectures for progressive refinement \cite{Li2020Reccurrent,yang2017high}, 
and involving the structural priors as guidance in a two-stage framework for structural consistency \cite{liao2018edge,nazeri2019edgeconnect,ren2019structureflow,Yang-structure}. However, these methods lack the ability to model long-term correlations between distant context, leading to blurry textures.

\begin{table}[tb]
    \centering
    \caption{\small Comparison of inpainting methods based on three dimensions: non-local correlation, structural guidance and framework.}
    \setlength\tabcolsep{9pt}
    \resizebox{0.9\linewidth}{!}{
    \begin{tabular}{lccc}
    \toprule
              &  Non-local  & Structures & Framework\\ \midrule
    GatedConv~\cite{yu2019free} & \cmark     & N/A & Two-steps \\
    PEN-Net~\cite{song2018spg}    & \cmark     & N/A & One-forward\\
    Edgeconnect~\cite{nazeri2019edgeconnect}    & \xmark     & Edge & Two-steps\\
    SPG-Net~\cite{song2018spg}    & \xmark     & Semantic & Two-steps\\    
    SGE-Net~\cite{Liao2020guidance}   & \xmark     & Semantic & Joint-learning\\
    Ours  & \cmark     & Semantic & Joint-learning       \\
    \bottomrule
    \end{tabular}
    }
    \label{tab:relatedwork}
\end{table}

To better refine the inpainted image textures, non-local algorithms are adopted to borrow distant features from a known region, which contains fine textures, to the missing region. \cite{yu2018generative} firstly proposed to compute textural affinity within the same image to ﬁll the corrupted area with more realistic texture patches from the available area. \cite{zeng2019learning} devised a pyramid of contextual attention at multiple layers to refine the textures from high-level to low-level. \cite{liu2019coherent} used a coherent semantic attention layer to ensure semantic relevance between nearby filled features. \cite{Xie2019Learnable} extends the single attention map to bidirectional attention maps and re-normalizes the features to let the decoder concentrate on filling the holes. Although these methods have delivered considerable improvements, they failed to address the semantic ambiguity as they try to measure the texture affinity across all semantics. 

\begin{figure*}[htb]
\centering
	\footnotesize{
		\begin{tabular}{c}   
            \includegraphics[width=\linewidth]{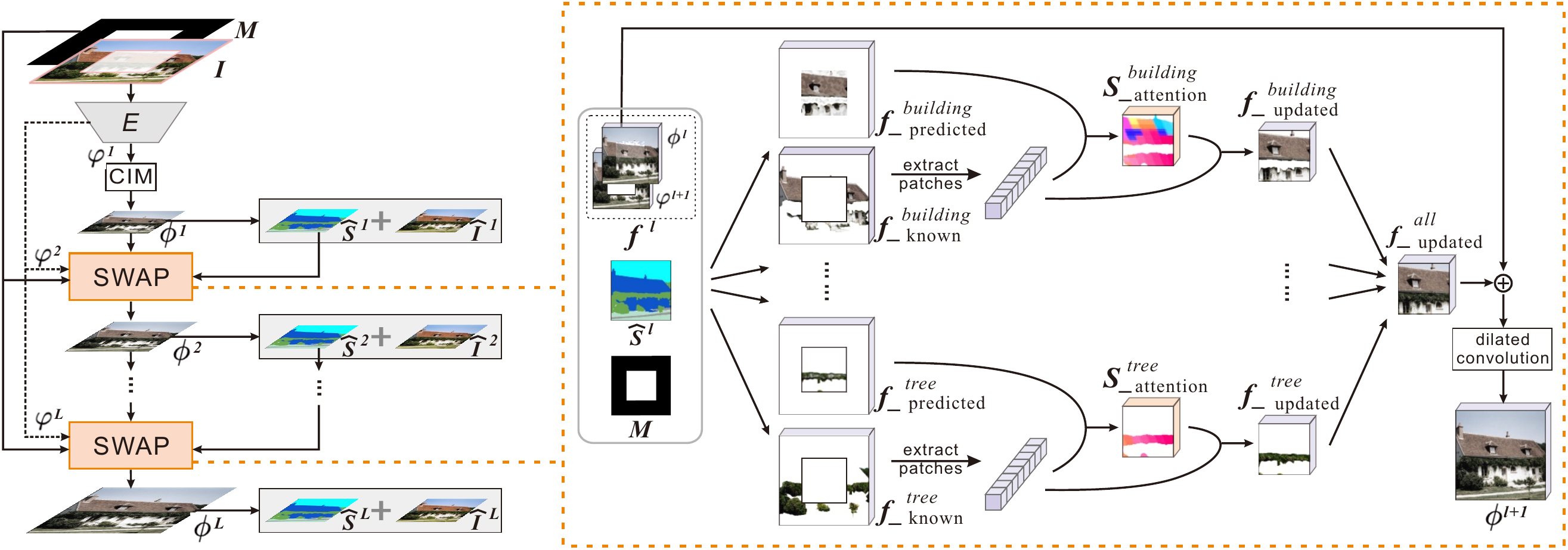} \\
			\small{(a) Overall Framework~~~~~~~~~~~~~~~~~~~~~~~~~~~~~~~~~~~~~~~~~~~(b) Semantic-wise Attention Propagation (SWAP) Module}\\
	\end{tabular}} 
\caption{\small Proposed network architecture. At each scale, both the inpainted image and segmentation map are output from two task-specific heads to control the predicted structures of the shared features. SWAP is added between scales to progressively optimize the texture details of contextual feature.}
\vspace{-1mm}
\label{fig:framework}
\end{figure*}

\subsection{Semantic-guided image processing}

Recent research reveals that the semantic priors of the high-level vision tasks are useful in the guidance of low-level vision tasks \cite{Chen2017photographic,Liu2020denoising,Chun2018synthesis}. One hot research topic is semantic-guided image generation. It covers several research directions, including image translation from segmentation maps to realistic images \cite{Isola2017translation,Wang2018perceptual} and semantic image synthesis \cite{Chen2017photographic,Tang2020local,Chun2018synthesis}. To avoid the vanishing of semantic priors in the generation process, \cite{park2019semantic} proposed a spatially adaptive normalization layer to propagate the semantic information to the synthesized images. With the useful semantic priors, it can generate high-quality images.

The semantic priors have also been applied to promote  many conditional low-level vision tasks and demonstrated its effectiveness in constraining the plausible solution space in the ill-posed problems. For example, they are involved in the tasks of super-resolution \cite{wang2018recovering}, dehazing \cite{Ren2019dehazing}, denoising \cite{Liu2020denoising}, style transfer \cite{Ma2020transfer,Tomei2019art2real} and image manipulation \cite{Hong2018manipulation,Ntavelis2020editing}. Inspired by the successful assistance from semantic priors in conditional image generation, we also exploit 
the semantic guidance in image inpainting.

\section{Proposed Method}
How to achieve high-quality inpainting results on both semantically reasonable structures and visually pleasing textures? We argue that such results should not only be able to reconstruct the structures of the semantic objects for the global structure, but also the textures of them should look realistic with same semantic in the image for local pixel continuity. To this end, we build a multi-task learning framework on the coherence priors to explicitly reconstruct both the structures and textures of semantic objects. Moreover, we propose a new SWAP module to optimize the textures by semantically binding the textures between an inpainted region and the known regions based on the coherence priors. we also devise two losses to guide the learning of coherence relationships both in the global structure level and in the local patch level, respectively.



\subsection{Framework Overview}
We build our network on an alternating-optimization architecture to utilize the coherence priors to mutually assist image inpainting and semantic segmentation for a corrupted image. Specially, we propose a multi-task learning framework by sharing features in the decoders for the two tasks (as shown in Figure~\ref{fig:framework}). The encoder encodes the corrupted image and its mask into hierarchical contextual features, which are then fed into the decoder to predict the inpainted images and semantic segmentation maps across scales. Prior to feeding the encoded feature of the last layer into the decoder, we initially complete the feature via a Context Inference Module (CIM) based on the contextual inference method \cite{Liao2020guidance,xiao2019cisi}. 

In the decoder,  the contextual feature at each scale is processed by two task-specific heads to predict the inpainted image and the segmentation map, respectively. Different from the method proposed in \cite{Liao2020guidance} that updates the contextual features by spatial adaptive normalization to capture the common properties of the same semantic, we propose a SWAP module to stress the realistic texture of each semantic patch by referring to the semantic relevant features from the known regions. In this way, the contextual features are learnt to represent the global structure and refined with the semantic-aware texture details.

For brevity, we adopt the following notations. $\varphi^l$ and $\phi^l$ denote the features from the encoder and from the decoder at scale $l$, respectively; $\hat{I}^l$ and $\hat{S}^l$ respectively represent the inpainted image and the predicted $k$-channel segmentation maps from the inpainting head $h(\cdot)$ and from the segmentation head $g(\cdot)$ at scale $l$, where $k$ is the total number of semantic labels; $l$ ranges from 1 (the coarsest layer) to 5 (the finest layer).

\subsection{Semantic-wise Attention Propagation (SWAP)}

The SWAP module is designed to optimize the contextual features by enhancing the semantic authenticity of textures based on coherence priors. As shown in Figure~\ref{fig:framework}(b), SWAP takes four inputs: two of them are the current-scale feature $\phi^l$ and the next-scale skip feature $\varphi^{l+1}$ from the encoder. The third is the predicted segmentation probability map $\hat{S}^l$, that is used to guide the separation of features. The last is the missing-region mask $M$. The propagation process can be formulated as follows:

\begin{equation}
 \centering
\phi^{l+1} = swap(\phi^{l}, \varphi^{l+1}, \hat{S}^l, M),
 \label{eq:infer_updating}
\end{equation}
where $swap(\cdot)$ is the process of refining the contextual features in SWAP.


\begin{figure}[tb]
\centering
	\footnotesize{
		\begin{tabular}{c}   
           \includegraphics[width=0.95\linewidth]{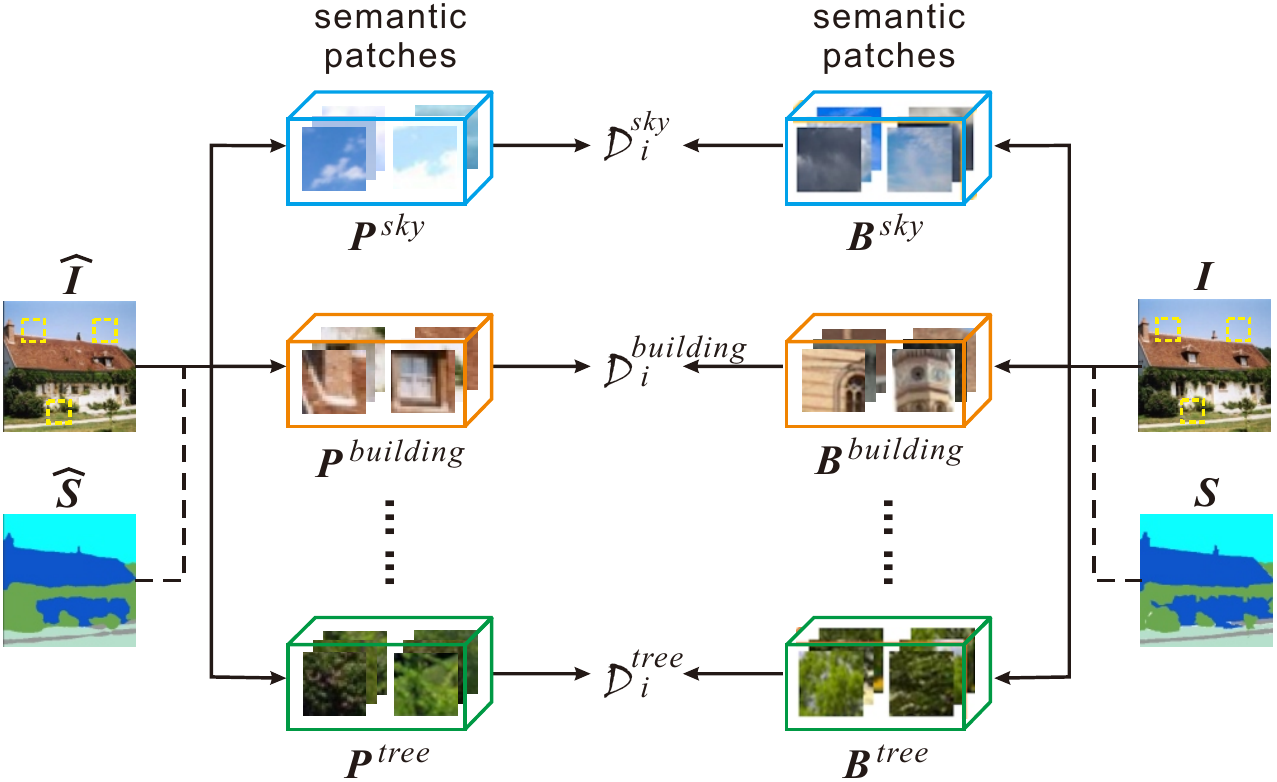}  \\
           	\small{(a) Non-local Patch Coherence Loss}\\
           \includegraphics[width=0.9\linewidth]{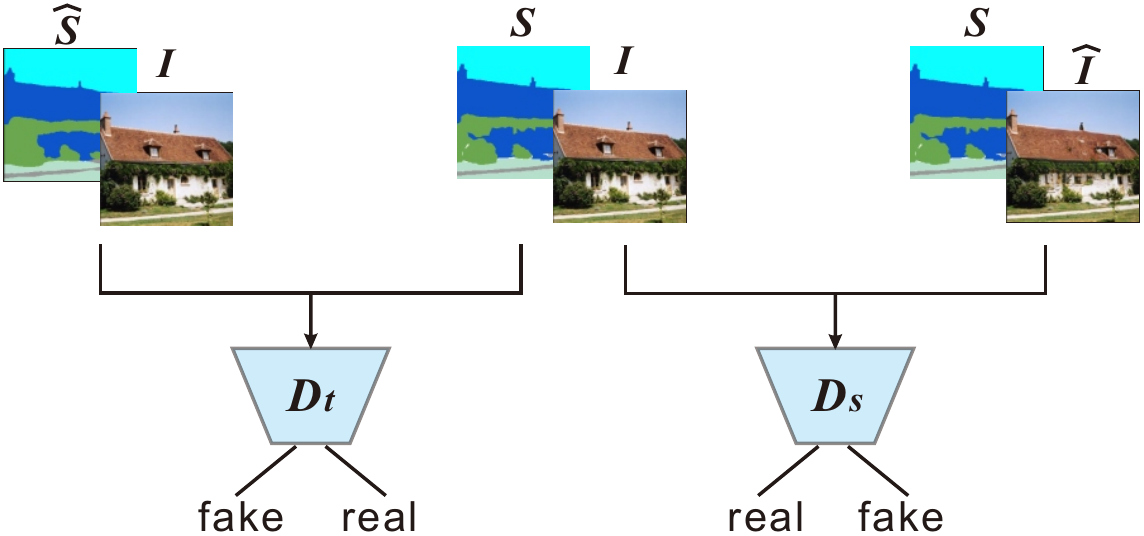}  \\
			\small{(b) Structure Coherence Loss}\\
	\end{tabular}} 
\caption{\small Proposed coherence losses. (a) non-local patch coherence loss encourages the generated texture to be as similar as any known patch of the same semantic in the real image; (b) structure coherence loss ensures the structural consistency between the entire segmentation map and the inpainted image.}
\vspace{-2mm}
\label{fig:loss}
\end{figure}

The attention-based approaches in \cite{song2018contextual,Xie2019Learnable,yu2018generative} resort to contextual attention to pick known regions as references to complete a missing region, which, however, cannot distinguish patches of different semantics and thereby leads to blurry boundaries and semantic confusion during attention propagation. Unlike these approaches, SWAP calculates the attention score by matching semantic-aware features of missing patches and the known patches based on the coherence priors. Specifically, we first split the contextual features $f^l$, which is generated from $\phi^l$ and $\varphi^{l+1}$, into different semantic parts according to the semantic labels in the predicted segmentation map $\hat{S}^l$ of the $l$-th layer. We subsequently drop the superscript $l$ for the sake of notational simplicity.

Within each semantic part, each patch's attention score is evaluated by the patch affinity between the missing region and a known region using the normalized inner product followed by a softmax operation:

\begin{equation}
    Df^c_{i,j} =\langle \frac{p^c_i}{\left|\left|p^c_i\right|\right|_2}, \frac{p^c_j}{\left|\left|p^c_j\right|\right|_2}\rangle,
\end{equation}

\begin{equation}
    \omega^c_{j,i} = \frac{\exp(Df^c_{i,j})}{\sum_{i=1}^N \exp(Df^c_{i,j})},
\end{equation}

\noindent where $p^c_i$ is the $i$-th patch extracted from semantic feature $f^c$ of class $c$ in the known region, $p^c_j$ is the $j$-th patch extracted from $f^c$ in the missing region. $Df^c_{i,j}$ is the affinity between them, and $\omega^c_{j,i}$ is the attention score representing the normalized affinities for each patch.

After obtaining the attention score from the known region, the feature of the $j$-th missing patch is updated by

\begin{equation}
    p^c_j = \sum_{i=1}^{N^c} \omega^c_{j,i}~p^c_i.
\end{equation}
\noindent where $N^c$ is the total number of patches of semantic class $c$ in the known region.

The output feature of SWAP is generated by merging the updated features of all semantics, followed by four groups of dilated convolutions with different rates to improve the structural coherence in the final reconstructed features.

\subsection{Coherence Losses}


We devise new coherence losses between the semantics and textures as supervisions to guide the image inpainting and semantic segmentation to meet the following requirements:
1) the overall structure between the inpainted image and segmentation map should match each other;
2) the predicted textures of a certain semantic class should have the same distribution as that of the semantic textures in the known region.  
Under these considerations, we propose two coherence losses shown in Figure~\ref{fig:loss}, a non-local patch coherence loss and a structure coherence loss, to respectively evaluate the patch similarity and the structural matching.

\heading{Non-local Patch Coherence Loss.}
Given the final inpainted image $\hat{I}$ and the predicted segmentation map $\hat{S}$, we aim to maximize the texture similarity between $\hat{I}$ and $I$. This is, the generated patches attributed to a specific class $c$ should be similar to the realistic patches of the same class in the ground-truth image.

Similar to the attention propagation process, we first split $\hat{I}$ and $I$ into different semantic images and extract patches to build the corresponding semantic patch sets $P^c=\{p_j^c\}$ and $B^c=\{b_i^c\}$ according to $\hat{S}$ and $S$. For each patch in $P^c$, we randomly select one patch from $B^c$ with the same semantic, and compute the pairwise cosine distances between them as

\begin{equation}
    Di^{c}_{i,j} = 1-\langle \frac{p^c_i-\mu_b^c}{\left|\left|p^c_i-\mu_b^c\right|\right|_2}, \frac{b^c_j-\mu_b^c}{\left|\left|b^c_j-\mu_b^c\right|\right|_2}\rangle,
\end{equation}

\noindent where $\mu_b^c=\frac{1}{N_c}\sum_j b_j^c$, $N_c$ is the number of patches in $B_c$. 

The non-local patch coherence loss for each semantic class $c$ aims to maximize the similarity between the patch couples:

\begin{equation}
    \mathcal{L}_{\text{nlc}}^c(P^c,B^c) = -\log(\frac{1}{N_P^c}(\sum_i~Di^{c}_{i,j})),
\end{equation}

\noindent where $N_P^c$ is the cardinality of the set of the generated patches with class label $c$. Our objective is defined as the sum of all the single-class non-local patch coherence losses over different classes found in $\hat{S}$:

\begin{equation}
    \mathcal{L}_{\text{nlco}}(I, \hat{I}, S, \hat{S}) = \sum_c~\mathcal{L}_{\text{nlco}}^c(P^c,B^c),
\end{equation}

\noindent where $c$ assumes all the class labels of mask in $\hat{S}$. Note that if the label value in $\hat{S}$ is not found in $S$, the coherence loss of the corresponding semantic patch set is set to 0.

\heading{Structure Coherence Loss.}
Besides the local patch similarity, we adopt a structure coherence loss to encourage the structural coherence between the inpainted image and the predicted segmentation map. In this work, we use two conditional discriminators to judge whether the semantics and the same image's textures are coherence. The texture-conditioned discriminator $D_t$ is introduced to detect the predicted segmentation map' ``fakes" given the real image, while the semantics-conditioned discriminator $D_s$ is trained to detect the inpainted image's ``fakes" given the real segmentation map. The structure coherence loss can be expressed as:

\begin{equation}
\begin{aligned}
\mathcal{L}_\text{sco}(I, \hat{I}, S, \hat{S})=\mathcal{L}_{c_s}(I, \hat{I}, S) + \mathcal{L}_{c_t}(S, \hat{S}, I)~~~~~~~~~~~~\\
~~~=\mathbb{E}_{I,S}[\log D_s(I,S)] +\mathbb{E}_{\hat{I},S}[\log(1- D_s(\hat{I},S))]\\
~~~+\mathbb{E}_{S,I}[\log D_t(S,I)] +\mathbb{E}_{\hat{S},I}[\log(1- D_t(\hat{S},I))].
\end{aligned}
 \label{eq:segs_gan}
\end{equation}

\subsection{Objective Functions}
We design appropriate supervised loss terms for learning
the inpainting and segmentation tasks at each scale to obtain
multi-scale predictions. We adopt the reconstruction loss, the adversarial loss and the proposed coherence losses to promote the fidelity of the inpainted images. The cross entropy loss is adopted for ensuring the accuracy of the predicted segmentation maps. 

\heading{Reconstruction Loss.} We use the $\mathcal{L}_1$ loss to encourage per-pixel reconstruction accuracy at all scales.
\begin{equation}
    \mathcal{L}_1(I,\hat{I}) = \sum_l~\left|\left|I-up(\hat{I}_l)\right|\right|.
\end{equation}
where $l$ is the scale. 5 scales are adopted in this work

\heading{Adversarial Loss.} We use a multi-scale PatchGAN \cite{Chun2018synthesis} to classify the global and local patches of an image at different resolutions. The discriminator at each scale is identical and only the input is a differently scaled version of an image.
\begin{equation}
\begin{split}
\mathcal{L}_\alpha(I, \hat{I}) = \sum_{k=1,2,3} (\mathbb{E}_{I}[\log D(p^{k}_{I})] + \mathbb{E}_{\hat{I}}[(1-\log D(p^k_{\hat{I}})]),
\end{split}
 \label{eq:adversarial_loss}
\end{equation}
where $D(\cdot)$ is the discriminator, $p^{k}_{I}$ and $p^k_{\hat{I}}$ are the patches in the $k$-th scaled versions of $I$ and $\hat{I}$.

\heading{Cross-Entropy Loss.} This loss is used to penalize the segmentation performance.
\begin{equation}
    \mathcal{L}_\text{xe}(S,\hat{S})=-\sum_l~\sum_{p\in S}S(p)\log(up(\hat{S}^l)(p)),
\end{equation}
where $p$ is the pixel index for segmentation map $S$.

\heading{Overall Training Loss.} The overall training loss function for our network is defined as the weighted sum of the above mentioned losses.
\begin{equation}
\begin{split}
\mathcal{L}_{Final}=\mathcal{L}_1(I, \hat{I})+\lambda_\alpha\mathcal{L}_{\alpha}(I, \hat{I})+\lambda_\text{xe}\mathcal{L}_\text{xe}(S,\hat{S}) \\ +\lambda_\text{co}(\mathcal{L}_\text{nlco}(I, \hat{I}, S, \hat{S})+\mathcal{L}_\text{sco}(I, \hat{I}, S, \hat{S})),
\end{split}
 \label{eq:training_loss}
\end{equation}
where $\lambda_\alpha$, $\lambda_{se}$ and $\lambda_\text{co}$ are the weights for the adversarial loss, cross-entropy loss and coherence loss, respectively.

\tabcolsep=0.5pt
\begin{figure*}[htb]
	\centering
  \linespread{0.8}
\footnotesize{
		\begin{tabular}{ccccccc}

			\includegraphics[width=0.13\textwidth]{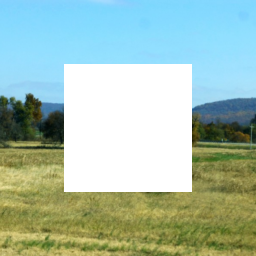} &
			\includegraphics[width=0.13\textwidth]{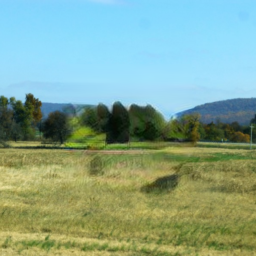} &
			\includegraphics[width=0.13\textwidth]{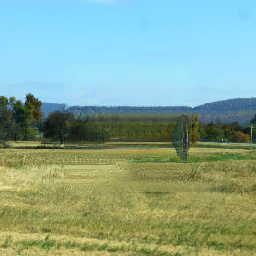} &
			\includegraphics[width=0.13\textwidth]{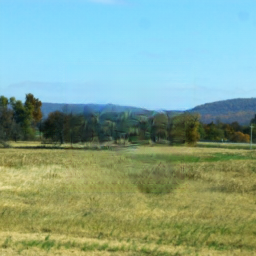} &
			\includegraphics[width=0.13\textwidth]{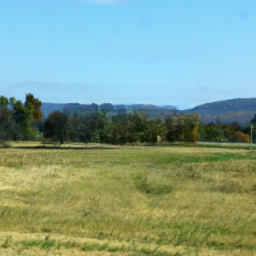} &
			\includegraphics[width=0.13\textwidth]{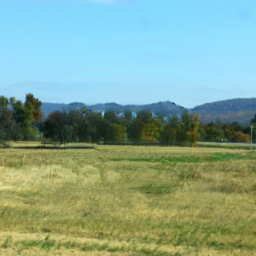} &
			\includegraphics[width=0.13\textwidth]{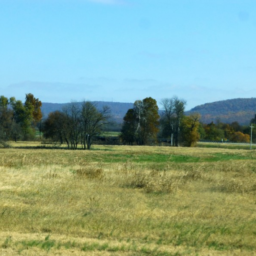} \\
			\includegraphics[width=0.13\textwidth]{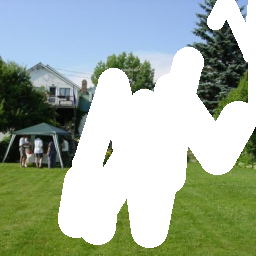} &
			\includegraphics[width=0.13\textwidth]{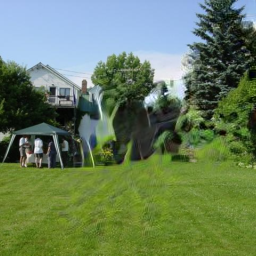} &
			\includegraphics[width=0.13\textwidth]{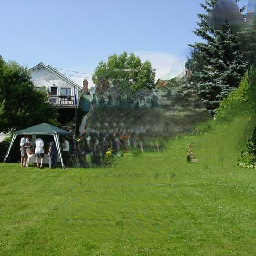} &
			\includegraphics[width=0.13\textwidth]{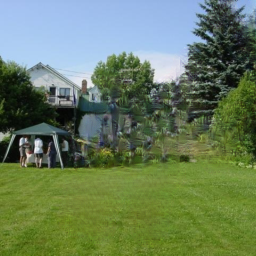} &
			\includegraphics[width=0.13\textwidth]{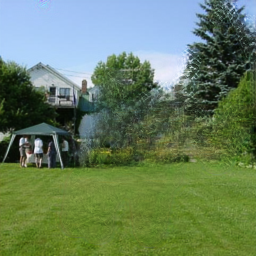} &
			\includegraphics[width=0.13\textwidth]{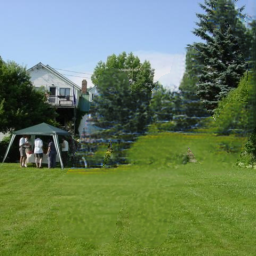} &
			\includegraphics[width=0.13\textwidth]{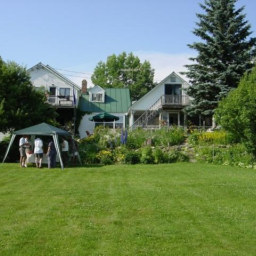} \\		\includegraphics[width=0.13\textwidth]{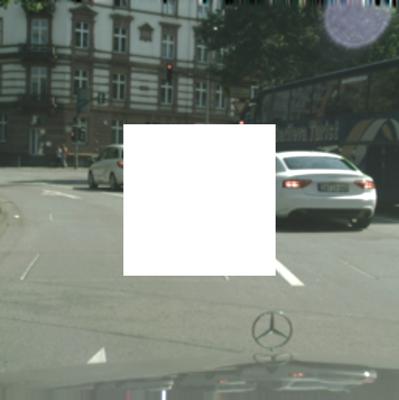}&
			\includegraphics[width=0.13\textwidth]{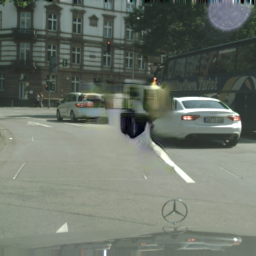} &
			\includegraphics[width=0.13\textwidth]{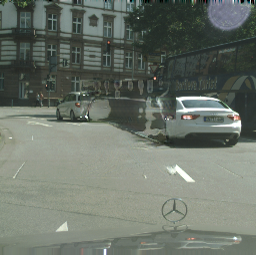} &
			\includegraphics[width=0.13\textwidth]{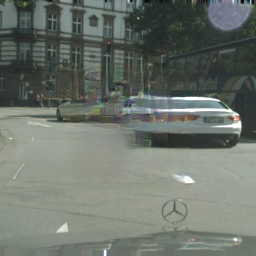} &
			\includegraphics[width=0.13\textwidth]{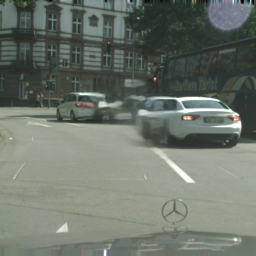} &
			\includegraphics[width=0.13\textwidth]{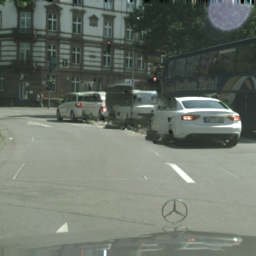} &
			\includegraphics[width=0.13\textwidth]{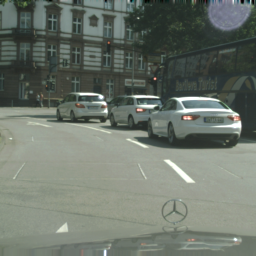} \\		\includegraphics[width=0.13\textwidth]{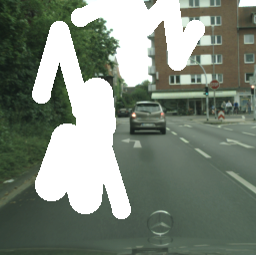}&
			\includegraphics[width=0.13\textwidth]{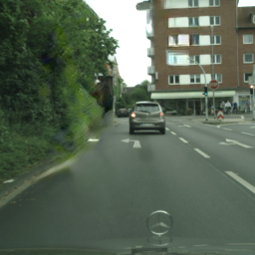} &
			\includegraphics[width=0.13\textwidth]{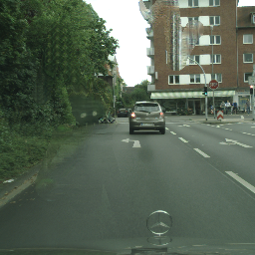} &
			\includegraphics[width=0.13\textwidth]{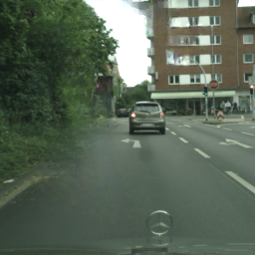} &
			\includegraphics[width=0.13\textwidth]{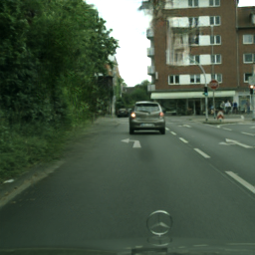} &
			\includegraphics[width=0.13\textwidth]{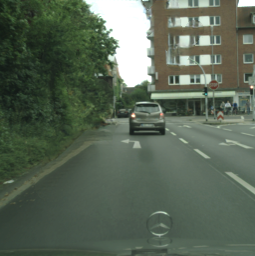} &
			\includegraphics[width=0.13\textwidth]{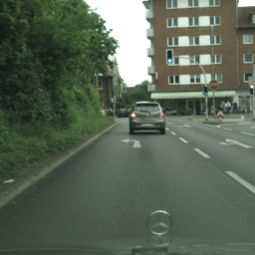} \\
			\small{(a) Input} & \small{(b) GatedConv} & \small{(c) PEN-Net} & \small{(d) EdgeConnect} & \small{(e) SGE-Net} & \small{(f) Our method} & \small{(g) Ground-truth} \\
	\end{tabular}}
   \caption{\small Qualitative comparison of inpainting results on image samples from \textbf{Outdoor Scenes} and  \textbf{Cityscapes}. Due to limited space, the comparison with SPG-Net is included in the supplementary material}
\label{fig:Overall-results}
\vspace{-2mm}
\end{figure*}

\section{Results}

\subsection{Experimental Settings}
We evaluate our method on the \textbf{Outdoor Scenes} \cite{wang2018recovering} and \textbf{Cityscapes} \cite{zhou2017places} datasets. \textbf{Outdoor Scenes} contains 9,900 training images and 300 test images. \textbf{Cityscapes} contains 5,000 street-view images in total. In order to enrich the training set of \textbf{Cityscapes}, we use 2,975 images from the training set and 1,525 images from the test set for training, and test on the 500 images from the validation set. Since the test set lacks human-labeled semantic annotations, we generate the annotations for training by using the state-of-the-art segmentation model Deeplab \cite{chen2018encoder}. We resize each training image to ensure its minimal height/width to be 256 for \textbf{Outdoor Scenes} and 512 for \textbf{Cityscapes}, and then randomly crop sub-images of size $256\times256$ as inputs to our model. The fine annotations of segmentation labels for both datasets are also provided for training, in which \textbf{Outdoor Scenes} and \textbf{Cityscapes} are annotated to 8 and 20 categories, respectively. Please note that the annotations can also be replaced by the extracted segmentation maps from the state-of-the-art segmentation models.

We compare our method with the following four learning-based inpainting methods: 1) GatedConv~\cite{yu2019free}: Contextual attention for leveraging the surrounding textures and structures. 2) EdgeConnect~\cite{nazeri2019edgeconnect}: Two-stage inpainting framework with edges as low-level structural information. 3) PEN-Net~\cite{zeng2019learning}: Cross-layer attention transfer and pyramid filling in a multi-scale framework. 4) SPG-Net~\cite{song2018spg}: Two-step inpainting with a semantic segmentation map as high-level structural information. 5) SGE-Net~\cite{Liao2020guidance}: Semantic guidance for inpainting based on spatially adaptive normalization \cite{park2019semantic}.


\begin{table*}[htb]
    \centering
    \caption{\small Objective quality comparison of six methods in terms of PSNR, SSIM, and FID on \textbf{Outdoor Scenes} and \textbf{Cityscapes} ($\uparrow$: Higher is better; $\downarrow$: Lower is better). The two best scores are colored in \rc{red} and \bc{blue}, respectively}
    \setlength\tabcolsep{7pt}
    \resizebox{0.95\linewidth}{!}{
    \begin{tabular}{l ccc ccc ccc ccc}
    \toprule
    & \multicolumn{6}{c}{Outdoor Scenes} & \multicolumn{6}{c}{Cityscapes} \\
    & \multicolumn{3}{c}{centering holes} & \multicolumn{3}{c}{irregular holes} & \multicolumn{3}{c}{centering holes} & \multicolumn{3}{c}{irregular holes} \\
    \cmidrule(lr){2-4} \cmidrule(lr){5-7} \cmidrule(lr){8-10} \cmidrule(lr){11-13}
    &PSNR$\uparrow$&SSIM$\uparrow$&FID$\downarrow$
    &PSNR$\uparrow$&SSIM$\uparrow$&FID$\downarrow$
    &PSNR$\uparrow$&SSIM$\uparrow$&FID$\downarrow$
    &PSNR$\uparrow$&SSIM$\uparrow$&FID$\downarrow$ \\ \midrule
    
GatedConv & 19.06& 0.73 & 42.34
                           & 18.47 & 0.74 & 44.15
                           & 21.13 & 0.74 & 20.03
                           & 17.13 & 0.67 & 43.14 \\
PEN-Net  & 18.58 & 0.75 & 44.12  
                           & 17.56 & 0.69 & 48.95
                           & 20.48 & 0.72 & 22.34
                           & 16.37 & 0.66 & 47.87 \\
EdgeConnect & 19.32 & \bc{0.76} & 41.25
                                        & 19.12 & 0.74 & 42.27
                                        & 21.71 & 0.76 & 19.87
                                        & 17.63 & 0.72 & \bc{39.04} \\
SPG-Net & 18.04 & 0.70 & 45.31  
                           & 17.85 &0.74 &50.03
                           & 20.14 & 0.71 & 23.21
                           & 16.41 &0.67 &43.63 \\
SGE-Net    & \bc{20.53} & \rc{0.81} & \bc{40.67}
                           & \bc{19.46} & 0.76 & \bc{39.14}
                           & \bc{23.41} & \rc{0.85} & \bc{18.67}
                           & \bc{17.78} & \bc{0.74} & 41.45 \\ 
\textbf{Ours} & \rc{21.18} & \rc{0.81} & \rc{38.15}
                           & \rc{20.31} & \rc{0.80} & \rc{36.74}
                           & \rc{23.89} & \bc{0.84} & \rc{18.14}
                           & \rc{17.86} & \rc{0.76} & \rc{38.18}  \\ 
    \bottomrule
    \end{tabular}
    }
    \label{tab:exp_stoa}
\end{table*}

\subsection{Qualitative Comparisons}
Figure~\ref{fig:Overall-results} shows the qualitative comparisons of our method with all the baselines. The corrupted area is simulated by sampling a central hole ($128\times128$ for \textbf{Outdoor Scenes} and $96\times96$ for \textbf{Cityscapes}) or randomly placing multiple irregular masks based on \cite{yu2019free}. As shown in the figure, the baselines usually suffer from artifacts and unsatisfactory boundaries while completing complex holes. GatedConv and PEN-Net adopt contextual attention to bring in the features of the known region, but they usually distort the structures when referencing to incorrect semantic textures from the surrounding, especially in completing the complex holes. EdgeConnect and SGE-Net are able to recover correct structures owning to the use of structure priors. However, EdgeConnect may generate mixed edges, making it difficult to generate correct textures, whereas the textures of SGE-Net are often over-smoothed without texture refinement. In contrast, our method generates more realistic textures and better boundaries delineating semantic regions than all the baselines thanks to the coherence priors between semantics and textures. 


\begin{figure}[tb]
    \linespread{0.5}
    	\centering
	\footnotesize{
		\begin{tabular}{ccccc}
			\includegraphics[width=0.2\columnwidth]{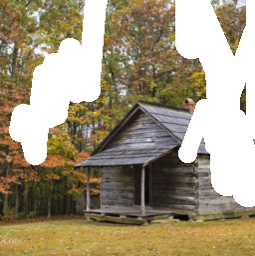} &
			\includegraphics[width=0.2\columnwidth]{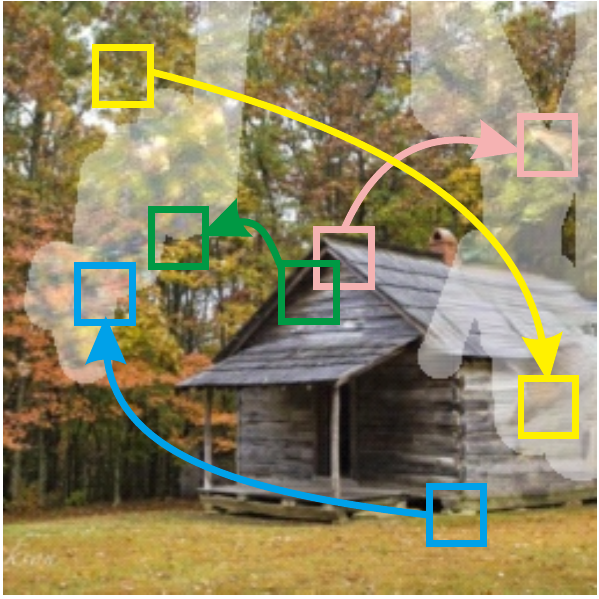} &
			\includegraphics[width=0.2\columnwidth]{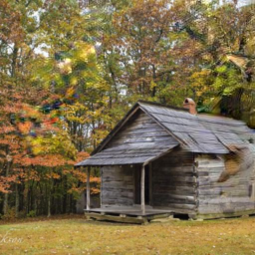} &
			\includegraphics[width=0.2\columnwidth]{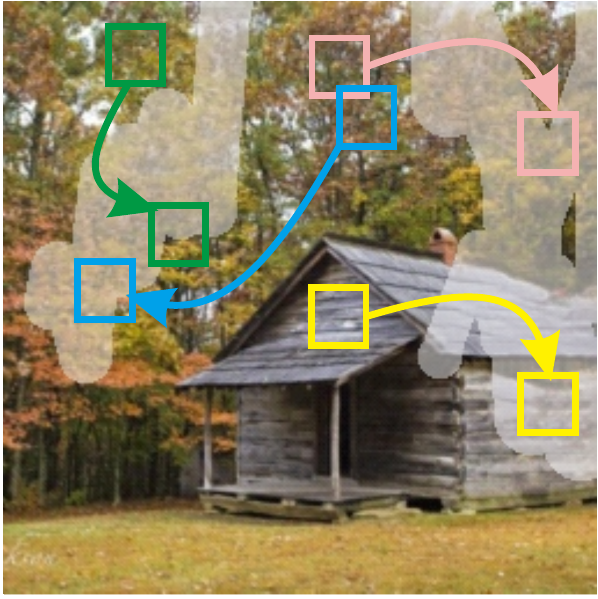} &
			\includegraphics[width=0.2\columnwidth]{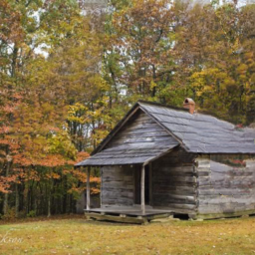}\\
			\includegraphics[width=0.2\columnwidth]{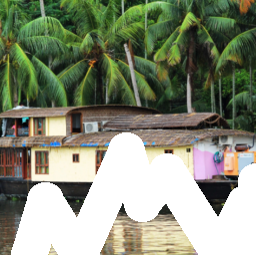} &
			\includegraphics[width=0.2\columnwidth]{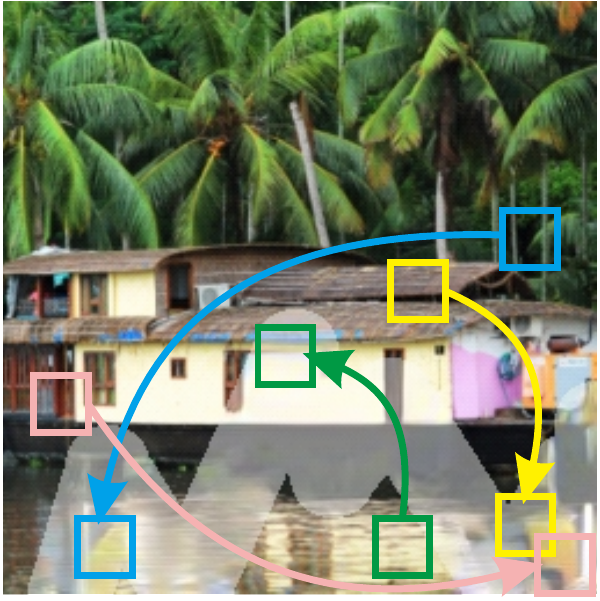} &
			\includegraphics[width=0.2\columnwidth]{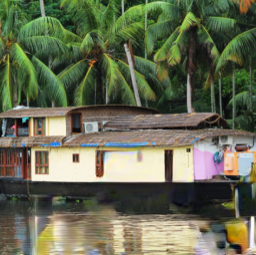} &
			\includegraphics[width=0.2\columnwidth]{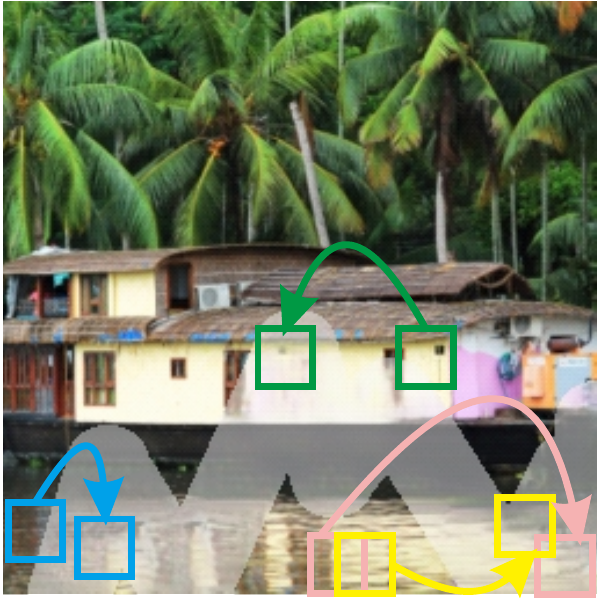} &
			\includegraphics[width=0.2\columnwidth]{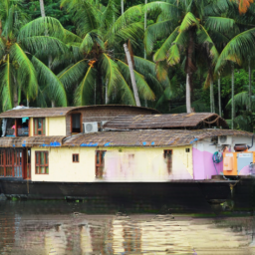}\\
	\end{tabular}} 
	    \linespread{1}
	\caption{\small Comparisons between different attention modules. From left to right: input, the most matched patches in existing attention module, result from existing attention module, the most matched patches in SWAP module, and result from SWAP module. The arrows in columns 2 and 4 indicate the matched patch from known region to the missing region.}
    \label{fig:SWAP}
\end{figure}


\begin{figure}[tb]
    \linespread{0.5}
    	\centering
	\footnotesize{
		\begin{tabular}{ccccc}
			\includegraphics[width=0.2\columnwidth]{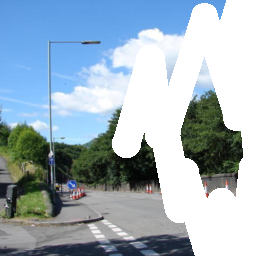} &
			\includegraphics[width=0.2\columnwidth]{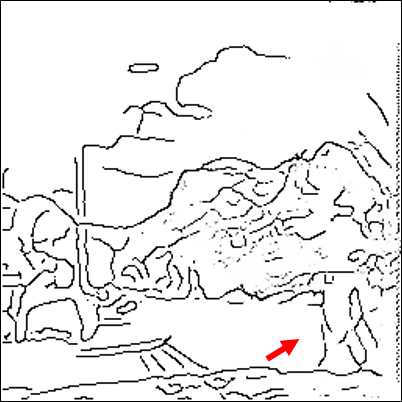} &
			\includegraphics[width=0.2\columnwidth]{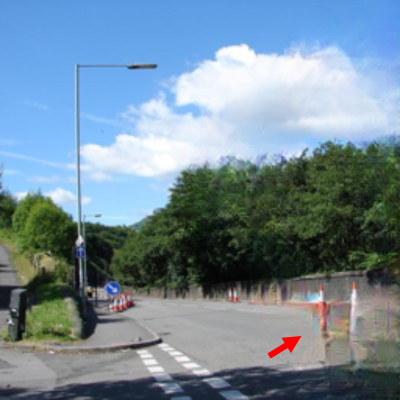} &
			\includegraphics[width=0.2\columnwidth]{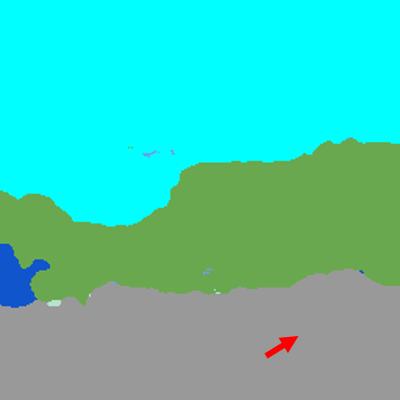} &
			\includegraphics[width=0.2\columnwidth]{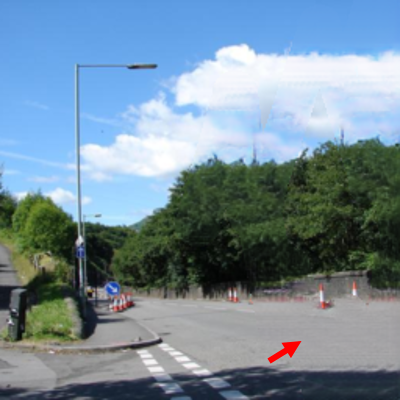} \\
			\includegraphics[width=0.2\columnwidth]{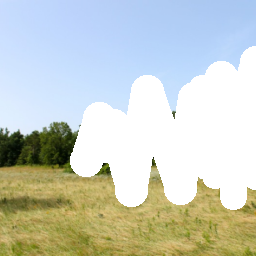} &
			\includegraphics[width=0.2\columnwidth]{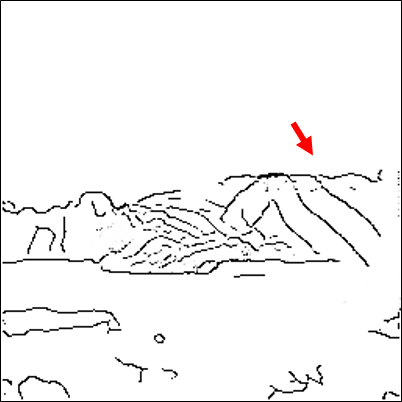} &
			\includegraphics[width=0.2\columnwidth]{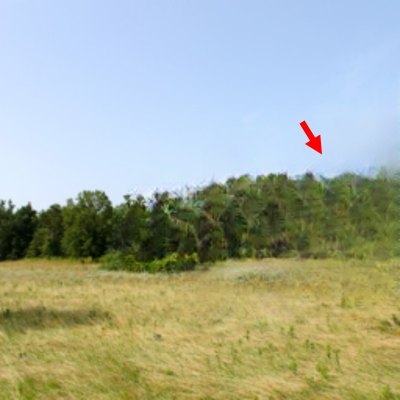} &
			\includegraphics[width=0.2\columnwidth]{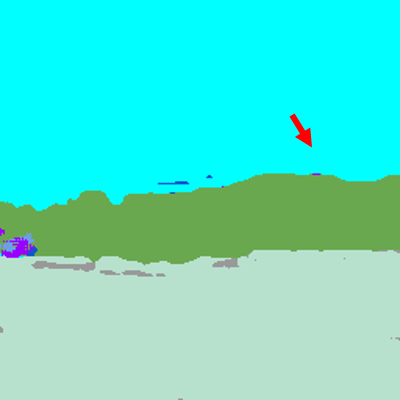} &
			\includegraphics[width=0.2\columnwidth]{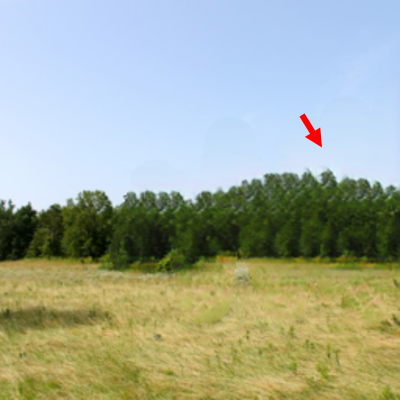} \\
	\end{tabular}}    
    \linespread{1}
	\caption{\small Visual quality comparisons between EdgeConnect and our method. From left to right: input, reconstructed edges and images by EdgeConnect, reconstructed segmentation maps and images by our method. The red arrows highlight the unrealistic regions generated from Edgeconnect compared with ours.}
    \label{fig:edge}
\vspace{-2mm}
\end{figure}

\subsection{Quantitative Comparisons}
Table~\ref{tab:exp_stoa} shows the quantitative comparisons on \textbf{Outdoor Scenes} and \textbf{Cityscapes} datasets based on three quality metrics: Peak Signal-to-Noise Ratio (PSNR), Structural Similarity Index (SSIM) and Fréchet Inception Distance (FID) \cite{heusel2017gans}. In general, the proposed method achieves significantly better objective scores than the baselines, especially in PSNR and SSIM.

\subsection{User Study}
We randomly select 100 images from the two datasets (50 from \textbf{Outdoor Scenes} and 50 from \textbf{Cityscapes}) and invite 20 subjects with image processing expertise to rank the subjective visual qualities of images inpainted by the five inpainting methods (GatedConv, PEN-Net, EdgeConnect, SPG-Net, SGE-Net, and our method). They are not informed of any mask information. For each test image, its five inpainting results are presented in a random order, and each subject is asked to rank the five methods from the best to the worst. The result shows that our method receives 51.8 \% favorite votes (i.e., the top-1 in 1,036 out of 2,000 comparisons), surpassing 19.3 \% with SEG-Net, 11.3 \% with EdgeConnect, 7.8 \% with GatedConv, 5.6 \% with PEN-Net, and 4.2 \% with SPG-Net. Hence, our method outperforms the other methods.

\subsection{Ablation Studies}
\subsubsection{Effectiveness of SWAP}
We verify the effectiveness of SWAP by comparing it with the contextual attention module from GatedConv \cite{yu2019free}. To show the difference, we highlight the location of the best-match patch for a patch in the missing area. As shown in Figure~\ref{fig:SWAP}, since the existing attention module refers to the whole known region without any semantic guide, it usually matches wrong texture patches to the missing area, leading to ambiguous textures. In contrast, benefiting from SWAP, our method matches the patches within the same semantic class, which effectively improves the fidelity of matched reference textures so as to generate more realistic textures. 

\begin{figure}[tb]
    \linespread{0.5}
    	\centering
	\footnotesize{
		\begin{tabular}{ccccc}
			\includegraphics[width=0.2\columnwidth]{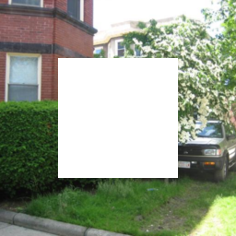} &
			\includegraphics[width=0.2\columnwidth]{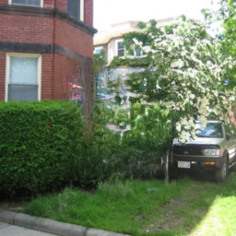} &
			\includegraphics[width=0.2\columnwidth]{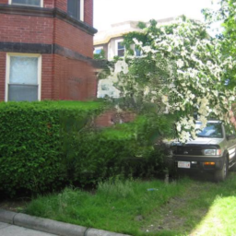} &
			\includegraphics[width=0.2\columnwidth]{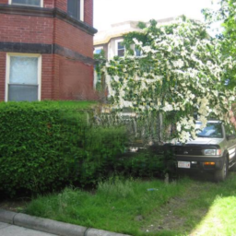} &
			\includegraphics[width=0.2\columnwidth]{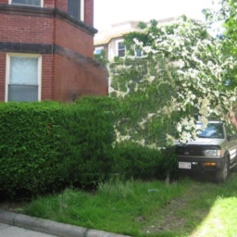} \\
			\includegraphics[width=0.2\columnwidth]{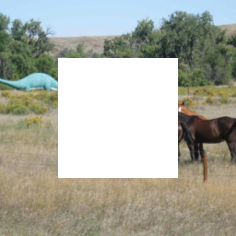} &
			\includegraphics[width=0.2\columnwidth]{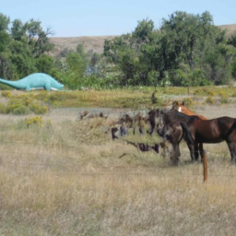} &
			\includegraphics[width=0.2\columnwidth]{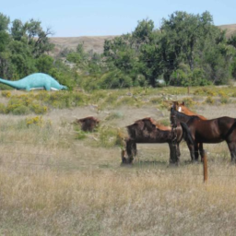} &
			\includegraphics[width=0.2\columnwidth]{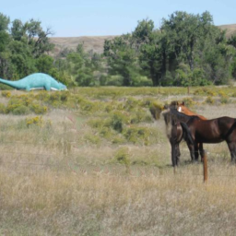} &
			\includegraphics[width=0.2\columnwidth]{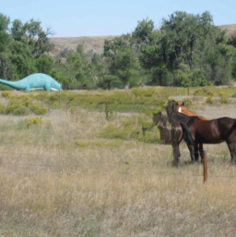} \\
	\end{tabular}}    
	    \linespread{1}
	\caption{\small Visual quality comparisons on four variants to show the effectiveness of SWAP and Coherence Loss. From left to right: input, results of Ours (Base), Ours (Att), Ours (SWAP), and Ours (Full).}
	\vspace{-4mm}
    \label{fig:SGIMandSEM}
\end{figure}

\tabcolsep=0.5pt
\begin{table}[tb]
    \centering
    \caption{\small Comparisons on the performance gains with SWAP and Coherence Loss (Co-Loss) in terms of three metrics.}
    \vspace{-2mm}
    \setlength\tabcolsep{7pt}
    \resizebox{0.95\linewidth}{!}{
    \begin{tabular}{lccccc}
    \toprule
              & SWAP & Co-Loss & PSNR$\uparrow$&SSIM$\uparrow$&FID$\downarrow$ \\ \midrule
    Ours (Base) & \xmark     & \xmark       &17.43 & 0.65 & 57.31 \\
    Ours (Att)    & \xmark     & \xmark       & 19.77 & 0.76 &43.54 \\
    Ours (SWAP)    & \cmark     & \xmark       & 20.58 & 0.79 &39.46 \\
    \textbf{Ours (Full)}   & \cmark     & \cmark       & \bf{21.18} & \bf{0.81} & \bf{38.15} \\
    \bottomrule
    \end{tabular}
    }
    \label{tab:SGIMandSEM}
\end{table}

\subsubsection{Edge \textit{vs.} Semantic Segmentation}
Our work assumes that semantic segmentation labels offer tighter clues to the textures than edges. To validate it, we compare the reconstructed structures of EdgeConnect and our model in Figure~\ref{fig:edge}. We find that the edges of different objects may be mixed up in the edge maps, making EdgeConnect fill in incorrect texture details for some missing areas. In contrast, the inferred semantic segmentation labels from our model help well delineate the layout of images, and the semantics can guide the filling of the textures, result in more photo-realistic results.

\subsubsection{Comparison of Segmentation Accuracy}

\begin{table}[tb]
\centering
    \setlength\tabcolsep{8pt}
\caption{\small Statistical comparison on semantic segmentation accuracy between the semantic-guided inpainting methods, namely SPG-Net \cite{song2018spg}, SGE-Net\cite{Liao2020guidance} and our method on \textbf{Outdoor Scenes} and \textbf{Cityscapes}.}
    \resizebox{0.75\linewidth}{!}{
    \begin{tabular}{lclc}
    \toprule
    \multicolumn{2}{c}{Outdoor Scenes} & \multicolumn{2}{c}{Cityscapes} \\
    \cmidrule(lr){1-2} \cmidrule(lr){3-4} 
    Methods & mIoU\% & Methods & mIoU\%   \\ \midrule
    \tabincell{l}{SPG-Net} & 0.51 & \tabincell{l}{SPG-Net}  & 0.39  \\ 
    SGE-Net     & 0.68 & SGE-Net   & 0.53  \\ 
    \textbf{Ours}    & \bf{0.71} & \textbf{Ours}  & \bf{0.57} \\ 
    \bottomrule
    \end{tabular}
    }
   \label{tab:segmentation-accuracy}
\end{table}


In order to further validate the coherence priors between semantics and the textures, We also conduct experiments to compare the generated segmentation maps from SPG-Net, SGE-Net and our method. Due to the alternative optimization of the inpainting and the segmentation tasks, we can generate high quality segmentation maps, which in turn improve the inpainting results. Table~\ref{tab:segmentation-accuracy} shows that our method outperforms the SPG-Net and the SGE-Net in semantic segmentation. The once-forward process from SPG-Net is hard to generate reliable semantic labels for large missing area, while the SGE-Net does not explicitly exploit the interaction between segmentation and inpainting. 



\subsubsection{Performance Gains with SWAP and Coherence Losses}
In our method, the two core components, SWAP and coherence loss, are devised to improve the inpainting performance. In order to investigate their effectiveness, we conduct an ablation study on four variants: a) Ours (Base), with only joint optimization of inpainting and segmentation in a multi-scale framework; b) Ours (Att), adopting the attention module \cite{yu2018generative} to measure the texture affinity across all semantics; c) Ours (SWAP), with SWAP; d) Ours (Full), with both SWAP and Coherence loss.

The visual and numeric comparisons on \textbf{Outdoor Scenes} are shown in Figure~\ref{fig:SGIMandSEM} and Table~\ref{tab:SGIMandSEM}, respectively. In general, the inpainting performance increases with the number of added modules. Specifically, the joint framework helps learn a more accurate scene layout, and the contextual attention does a good job of generating detailed content. Our SWAP can identify more relevant textures thanks to the predicted semantics. Moreover, the coherence losses further improves the texture details of inpainted regions.

\begin{figure}[tb]
    \linespread{0.5}
    	\centering
	\footnotesize{
		\begin{tabular}{ccccc}
			\includegraphics[width=0.2\columnwidth]{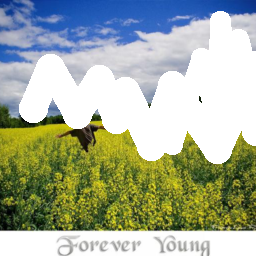} &
			\includegraphics[width=0.2\columnwidth]{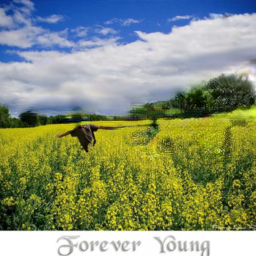} &
			\includegraphics[width=0.2\columnwidth]{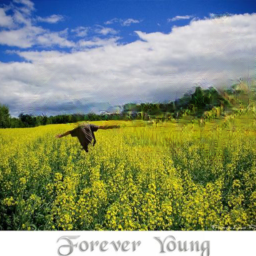} &
			\includegraphics[width=0.2\columnwidth]{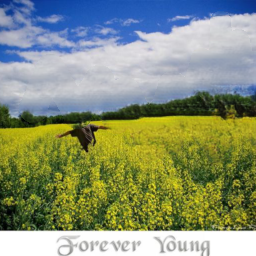} &
			\includegraphics[width=0.2\columnwidth]{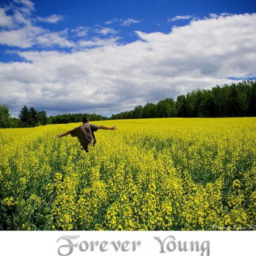} \\
			\includegraphics[width=0.2\columnwidth]{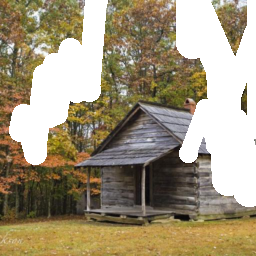} &
			\includegraphics[width=0.2\columnwidth]{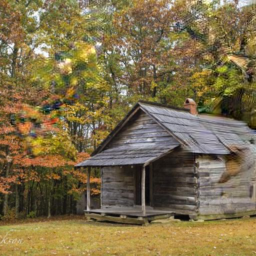} &
			\includegraphics[width=0.2\columnwidth]{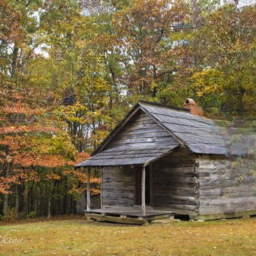} &
			\includegraphics[width=0.2\columnwidth]{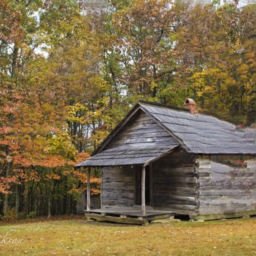} &
			\includegraphics[width=0.2\columnwidth]{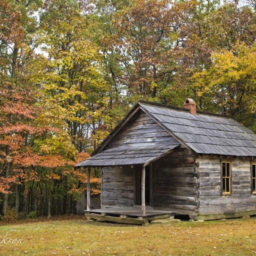} \\
	\end{tabular}}  
	    \linespread{1}
	\caption{\small Visual quality comparisons on image samples from \textbf{Places2}. From left to right: input, results of GatedConv, EdgeConnect, and Our method, ground-truth.}
    \label{fig:Places2}
\end{figure}

\subsubsection{Additional Results on Places2}
We also conduct performance evaluation on the \textbf{Places2} dataset \cite{zhou2017places} without semantic annotation, which was used in the assessment by both GatedConv and EdgeConnect. We use our model trained on \textbf{Outdoor Scenes} to complete the images with similar scenes in \textbf{Places2}. The subjective results in Figure~\ref{fig:Places2} show that our model is still able to generate proper semantic structures and textures, owing to the supervision of the coherence loss, which provides better prior knowledge about the scenes. 

\section{Conclusion}
We proposed a novel joint optimization framework of semantic segmentation and image inpainting to exploit the coherence priors existed between semantics and textures for solving the complex holes inpainting problem. To address the irrelevant texture filling, we proposed a semantic-wise attention propagation module to optimize the predicted textures from the same semantical region and two coherence losses to constrain the consistency of the semantic and texture in the same image. Experimental results demonstrate that our method can effectively generate promising semantic structures and texture details.

{\small

\bibliographystyle{ieee_fullname}
}

\end{document}